\documentclass{article}

    \usepackage[preprint]{neurips_2024}

\usepackage[utf8]{inputenc} 
\usepackage[T1]{fontenc}    
\usepackage{url}            
\usepackage{booktabs}       
\usepackage{amsfonts}       
\usepackage{nicefrac}       
\usepackage{microtype}      
\usepackage{xcolor}         

\usepackage{amsmath}
\usepackage{amssymb}
\usepackage{mathtools}
\usepackage{amsthm}

\usepackage{caption}
\usepackage{subcaption}

\usepackage{multirow}
\usepackage{comment} 
\usepackage{enumitem}

\usepackage{times}
\usepackage{epsfig}
\usepackage{graphicx}
\usepackage{wrapfig}

\usepackage[bb=dsserif]{mathalpha}
\usepackage{bm}

\usepackage{booktabs,colortbl,tabularx}
\usepackage{pifont}%

\definecolor{Gray}{gray}{0.9}
\definecolor{citecolor}{HTML}{2980b9}
\definecolor{linkcolor}{HTML}{c0392b}

\usepackage[pagebackref=true,breaklinks=true,colorlinks,bookmarks=false,citecolor=blue,linkcolor=red]{hyperref}

\usepackage[capitalize,noabbrev]{cleveref}

\definecolor{battleshipgrey}{rgb}{0.52, 0.52, 0.51}

\title{Efficient 3D Shape Generation via Diffusion Mamba with Bidirectional SSMs}

\author{%
  Shentong Mo\thanks{Corresponding author: shentongmo@gmail.com}\\
  Carnegie Mellon University
}

\begin{document}

\maketitle

\begin{abstract}

Recent advancements in sequence modeling have led to the development of the Mamba architecture, noted for its selective state space approach, offering a promising avenue for efficient long sequence handling. However, its application in 3D shape generation, particularly at high resolutions, remains underexplored. Traditional diffusion transformers (DiT) with self-attention mechanisms, despite their potential, face scalability challenges due to the cubic complexity of attention operations as input length increases. This complexity becomes a significant hurdle when dealing with high-resolution voxel sizes. To address this challenge, we introduce a novel diffusion architecture tailored for 3D point clouds generation—Diffusion Mamba (DiM-3D). This architecture forgoes traditional attention mechanisms, instead utilizing the inherent efficiency of the Mamba architecture to maintain linear complexity with respect to sequence length. DiM-3D is characterized by fast inference times and substantially lower computational demands, quantified in reduced Gflops, thereby addressing the key scalability issues of prior models. Our empirical results on the ShapeNet benchmark demonstrate that DiM-3D achieves state-of-the-art performance in generating high-fidelity and diverse 3D shapes. Additionally, DiM-3D shows superior capabilities in tasks like 3D point cloud completion. This not only proves the model's scalability but also underscores its efficiency in generating detailed, high-resolution voxels necessary for advanced 3D shape modeling, particularly excelling in environments requiring high-resolution voxel sizes. Through these findings, we illustrate the exceptional scalability and efficiency of the Diffusion Mamba framework in 3D shape generation, setting a new standard for the field and paving the way for future explorations in high-resolution 3D modeling technologies.

\end{abstract}

\section{Introduction}

Efficient and scalable image generation models~\cite{ho2020denoising,song2021scorebased,song2021denoisingdi} are a central pursuit in the field of machine learning, particularly within the subdomain of generative modeling. 
Recent years have seen significant strides in this area, with diffusion models~\cite{ho2020denoising,song2021scorebased} and self-attention transformers~\cite{Peebles2022DiT,bao2022all,bao2023transformer} at the forefront of innovation.
The burgeoning field of 3D shape generation~\cite{zhou2021pvd,zeng2022lion,gao2022get3d,liu2023meshdiffusion} is a cornerstone of numerous applications, ranging from augmented reality and virtual environments to industrial design and autonomous navigation.

However, as the demand for higher resolution and more complex shapes increases, existing methodologies struggle to keep pace, particularly when scaling to larger sequence lengths or higher voxel resolutions. Traditional approaches, such as diffusion transformers (DiT-3D)~\cite{mo2023dit3d,mo2023fastdit3d} that leverage self-attention mechanisms, although promising, are hindered by the cubic complexity of attention operations relative to input length. 
This complexity barrier poses significant challenges, particularly in resource-intensive scenarios involving high-resolution 3D shape generation.

Recent advancements in sequence modeling have introduced architectures like Mamba~\cite{gu2023mamba}, which utilizes a selective state space approach to efficiently manage long sequences. Despite its notable successes in other domains, the Mamba architecture's potential in 3D shape generation, especially at high resolutions, remains largely untapped. Recognizing this, our work explores the adaptation of the Mamba framework to overcome the limitations posed by traditional diffusion models in 3D shape generation.

In this paper, we present Diffusion Mamba for 3D shape generation, namely DiM-3D, a novel architecture designed specifically for the generation of 3D point clouds. DiM-3D eschews traditional attention-based mechanisms in favor of a more scalable approach, leveraging the efficiency and scalability inherent to the Mamba architecture. By doing so, it maintains linear complexity in relation to sequence length, enabling rapid inference and reduced computational demands measured in lower Gflops. This design not only addresses the critical scalability issues faced by prior models but also enhances performance in generating detailed, high-resolution voxel outputs.

Our contributions are substantiated through extensive experiments on the ShapeNet benchmark, where DiM-3D demonstrates state-of-the-art performance in generating high-fidelity and diverse 3D shapes. Furthermore, it shows remarkable results in 3D point cloud completion tasks. The capabilities of DiM-3D underline a significant step forward in the field, showcasing the model's efficiency and scalability, especially notable in handling 256-resolution voxel sizes.
Through our research, we aim to set a new benchmark for 3D shape generation, providing a robust framework that not only meets but exceeds the current demands of the industry. This paper details the architecture of DiM-3D, discusses its implementation, and evaluates its performance against existing standards, thereby illustrating its potential as a transformative tool in the realm of high-resolution 3D modeling.

Our main contributions can be summarized as follows:
\begin{itemize}
    \item We introduce DiM-3D, a novel diffusion mamba architecture for 3D shape generation that leverages the efficiency of state space models (SSMs) to address the computational and scalability challenges inherent in high-resolution 3D point cloud generation.
    \item Through extensive experiments on the ShapeNet dataset, DiM-3D demonstrates superior performance in generating high-fidelity and diverse 3D shapes compared to existing methods.
\item We provide a comprehensive analysis of DiM-3D's scalability and efficiency, showcasing its capability to handle larger datasets and diverse classes with various model sizes.
\end{itemize}

\section{Related Work}

\noindent\textbf{State Space Models.}
State space models (SSMs) have experienced a resurgence in the field of machine learning due to their efficiency in processing long sequences with fewer parameters and reduced computational demands~\cite{gu2023mamba,fu2023hungry,gu2022efficiently}. Notable implementations such as S4~\cite{gu2022efficiently} and Mamba~\cite{gu2023mamba} by Gu et al. demonstrate the capability of SSMs to outperform traditional recurrent neural networks in terms of scalability and efficiency . In our work, we extend these principles to the domain of 3D point cloud generation, harnessing the linear computational complexity of SSMs to significantly enhance the scalability and performance of generative tasks.

\noindent\textbf{Diffusion Models.}
Diffusion models~\cite{ho2020denoising,song2021scorebased,song2021denoisingdi} have emerged as powerful tools in generative tasks across various domains, including image generation~\cite{saharia2022photorealistic,mo2024scaling}, image restoration~\cite{saharia2021image}, audio generation~\cite{kong2021diffwave,mo2024texttoaudio,zhang2024audiosynchronized}, and video generation~\cite{ho2022imagen}.
Denoising Diffusion Probabilistic Models (DDPMs)~\cite{ho2020denoising,song2021scorebased} and their variants, such as DDIM, employ a forward process that incrementally introduces Gaussian noise, coupled with a reverse process trained to recover the original data from the noised versions . Our approach builds on the foundational principles of DDPMs but adapts them to the complexities of generating 3D point clouds, thereby advancing the application of diffusion models.

\noindent\textbf{Diffusion Transformers.}
Diffusion Transformers have recently shown their impressive capacity to generate high-fidelity images~\cite{Peebles2022DiT,bao2022all,bao2023transformer} and point clouds~\cite{mo2023dit3d,mo2023fastdit3d}. 
Typically, Diffusion Transformer (DiT)~\cite{Peebles2022DiT} proposed a plain diffusion Transformer architecture to learn the denoising diffusion process on latent patches from a pre-trained pre-trained variational autoencoder model in Stable Diffusion~\cite{Rombach2022highresolution}.
U-ViT~\cite{bao2022all} incorporated all the time, condition, and noisy image patches as tokens
and utilized a Vision transformer(ViT)~\cite{Dosovitskiy2021vit}-based architecture with long skip connections between shallow and deep layers.
More recently, UniDiffuser~\cite{bao2023transformer} designed a unified transformer for diffusion models to handle input types of different modalities by learning all distributions simultaneously.

The recent advent of Diffusion Transformers (DiTs)~\cite{Peebles2022DiT,bao2022all,bao2023transformer} has significantly advanced the field of high-fidelity generative modeling. Models like the original DiT and subsequent developments such as U-ViT and UniDiffuser showcase the integration of transformer architectures with diffusion principles to enhance generative capabilities across different data modalities. DiT-3D~\cite{mo2023dit3d}, for instance, specifically adapts the DiT architecture for 3D point cloud generation, demonstrating impressive results in handling complex 3D data.
In the context of our work, we introduce DiM-3D, a novel diffusion mamba framework that incorporates the efficiency of Mamba-based SSMs with the generative prowess of diffusion transformers. This integration allows DiM-3D to achieve state-of-the-art performance in generating high-fidelity and diverse 3D shapes while maintaining manageable computational demands. Our model not only advances the technological capabilities of 3D generative modeling but also contributes to the broader understanding and application of efficient generative models in various industrial and creative domains.

\begin{figure*}[t]
\centering
\includegraphics[width=0.99\linewidth]{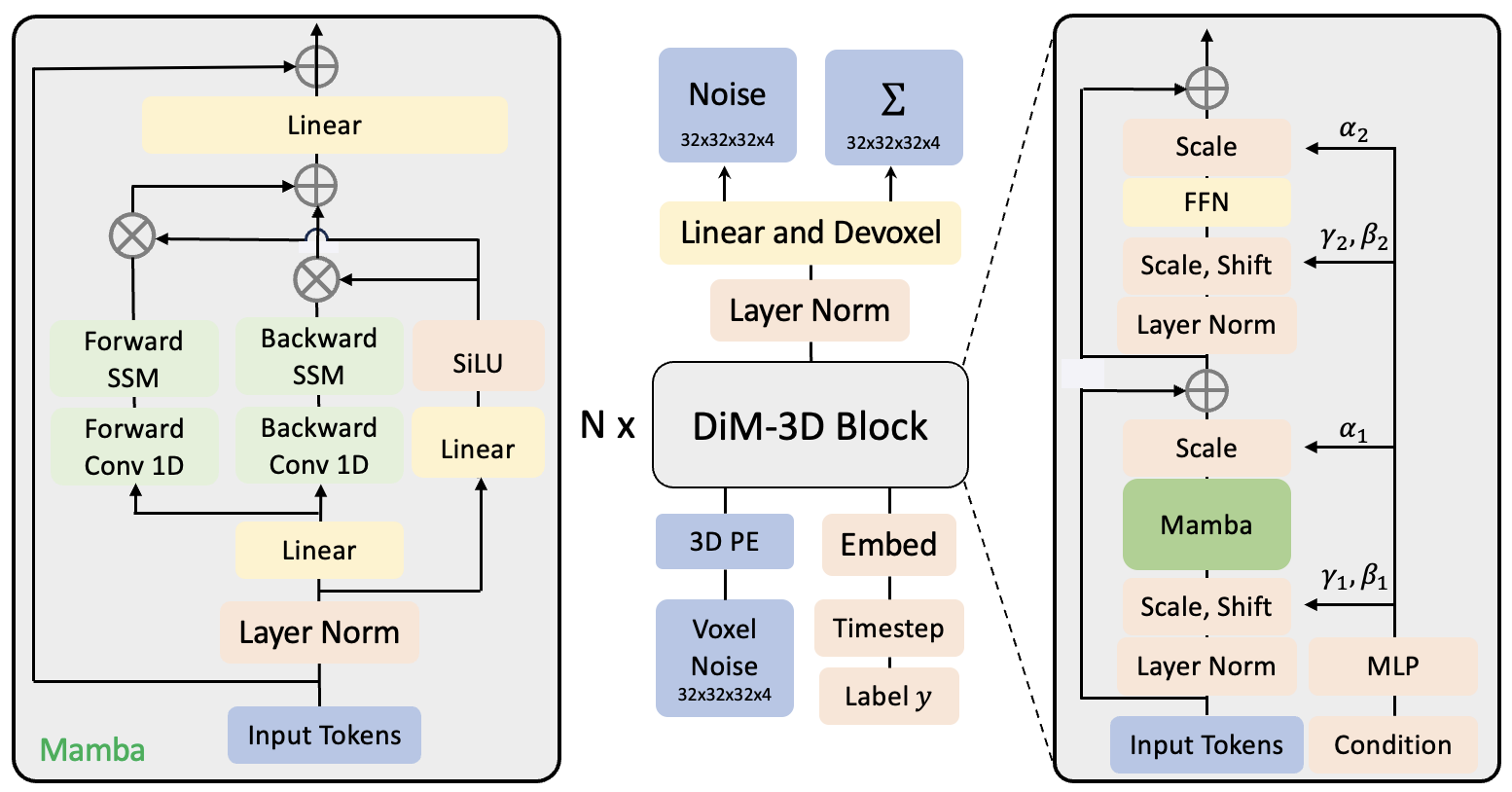}
\caption{{\bf Illustration of the proposed Diffusion Mamba framework for 3D shape generation (DiM-3D).}
The framework takes voxelized point clouds as input, and a patchification operator is used to generate token-level patch embeddings.
Then, multiple DiM-3D blocks based on Mamba with bidirectional state space models (SSMs) extract representations from all input tokens.
Finally, a linear layer and a devoxelization operator are used to predict the noise in the point cloud space.
}
\label{fig: main_img}
\end{figure*}

\section{Method}

In this paper, we present DiM-3D, a novel diffusion mamba architecture tailored for generating high-fidelity 3D point clouds. This architecture effectively leverages the selective state space approach of the Mamba framework to handle the inherent complexities of 3D shape generation. We introduce two main modules within this architecture: Diffusion Mamba for 3D shape generation (DiM) in Section~\ref{sec:diffmam}, and the DiM-3D block in Section~\ref{sec:dim}, which maintains linear complexity and facilitates efficient scaling to high resolutions.

\subsection{Preliminaries}

In this section, we first describe the problem setup and notations and then revisit DDPMs and diffusion transformers for image generation and SSMs for long sequence modeling.

\noindent\textbf{Problem Setup and Notations.}
We begin with a dataset of 3D shapes, represented as point clouds, $\mathcal{S} = {\mathbf{p}_i}_{i=1}^S$, where each point cloud $\mathbf{p}_i$ comprises $N$ points in $\mathbb{R}^{N\times 3}$. These point clouds are categorized into $M$ distinct classes, with each shape $\mathbf{p}_i$ associated with a class label ${y_i}^M_{i=1}$. Our objective is to train a diffusion model, DiM-3D, to generate new, high-fidelity point clouds. The model employs classifier-free guidance in its conditional diffusion process, similar to prior diffusion transformers for images.

\noindent\textbf{Revisit DDPMs.} 
The 3D shape generation problem, previous work~\cite{zhou2021pvd} based on denoising diffusion probabilistic models (DDPMs) define a forward noising process that gradually applies noise to real data $\mathbf{x}_0$ as $q(\mathbf{x}_t|\mathbf{x}_{t-1}) = \mathcal{N}(\mathbf{x}_t; \sqrt{1-\beta_t}\mathbf{x}_{t-1}, \beta_t\mathbf{I})$, where $\beta_t$ is a Gaussian noise value between $0$ and $1$.
In particular, the denoising process produces a series of shape variables with decreasing levels of noise, denoted as $\mathbf{x}_T, \mathbf{x}_{T-1}, ..., \mathbf{x}_0$, where $\mathbf{x}_T$ is sampled from a Gaussian prior and $\mathbf{x}_0$ is the final output.
With the reparameterization trick, we can have $\mathbf{x}_t = \sqrt{\bar{\alpha}_t}\mathbf{x}_0 + \sqrt{1-\bar{\alpha}_t}\boldsymbol{\epsilon}$, where $\boldsymbol{\epsilon}\sim \mathcal{N}(\mathbf{0},\mathbf{I})$, $\alpha_t = 1-\beta_t$, and $\bar{\alpha}_t = \prod_{i=1}^t \alpha_i$.

For the reverse process, diffusion models are trained to learn a denoising network $\boldsymbol{\theta}$ for inverting forward process corruption as $p_{\boldsymbol{\theta}}(\mathbf{x}_{t-1}|\mathbf{x}_t) = \mathcal{N}(\mathbf{x}_{t-1}; \boldsymbol{\mu}_{\boldsymbol{\theta}}(\mathbf{x}_{t},t), \sigma_t^2\mathbf{I})$.
The training objective is to maximize a variational lower bound of the negative log data likelihood that involves all of $\mathbf{x}_0, ..., \mathbf{x}_T$ as
\begin{equation}
\begin{aligned}
    \mathcal{L} = -p_{\boldsymbol{\theta}}(\mathbf{x}_{0}|\mathbf{x}_1) + \sum_{t} \mathcal{D}_{\text{KL}}(q(\mathbf{x}_{t-1}|\mathbf{x}_{t}, \mathbf{x}_{0})||p_{\boldsymbol{\theta}}(\mathbf{x}_{t-1}|\mathbf{x}_t)))
\end{aligned}
\end{equation}
where $\mathcal{D}_{\text{KL}}(\cdot||\cdot)$ denotes the KL divergence measuring the distance between two distributions.
Since both $p_{\boldsymbol{\theta}}(\mathbf{x}_{t-1}|\mathbf{x}_t))$ and $q(\mathbf{x}_{t-1}|\mathbf{x}_{t}, \mathbf{x}_{0})$ are Gaussians, we can reparameterize $\boldsymbol{\mu}_{\boldsymbol{\theta}}(\mathbf{x}_{t},t)$ to predict the noise $\boldsymbol{\epsilon}_{\boldsymbol{\theta}}(\mathbf{x}_{t},t)$.
In the end, the training objective can be reduced to a simple mean-squared loss between the model output $\boldsymbol{\epsilon}_{\boldsymbol{\theta}}(\mathbf{x}_{t},t)$ and the ground truth Gaussian noise $\boldsymbol{\epsilon}$ as:
$\mathcal{L}_{\text{simple}} = \|\boldsymbol{\epsilon}-\boldsymbol{\epsilon}_{\boldsymbol{\theta}}(\mathbf{x}_{t},t)\|^2$.
After $p_{\boldsymbol{\theta}}(\mathbf{x}_{t-1}|\mathbf{x}_t))$ is trained, new point clouds can be generated by progressively sampling $\mathbf{x}_{t-1}\sim p_{\boldsymbol{\theta}}(\mathbf{x}_{t-1}|\mathbf{x}_t))$ by using the reparameterization trick with initialization of $\mathbf{x}_{T}\sim \mathcal{N}(\mathbf{0},\mathbf{I})$.

\noindent\textbf{Revisit Diffusion Transformer (DiT).}
To generate high-fidelity 2D images, DiT proposed to train latent diffusion models (LDMs) with Transformers as the backbone, consisting of two training models.
They first extract the latent code $\mathbf{z}$ from an image sample $\mathbf{x}$ using an autoencoder with an encoder $f_{\text{enc}}(\cdot)$ and a decoder $f_{\text{dec}}(\cdot)$, that is, $\mathbf{z} = f_{\text{enc}}(\mathbf{x})$. 
The decoder is used to reconstruct the image sample $\hat{\mathbf{x}}$ from the latent code $\mathbf{z}$, \textit{i.e.},  $\hat{\mathbf{x}} = f_{\text{dec}}(\mathbf{z})$.
Based on latent codes $\mathbf{z}$, a latent diffusion transformer with multiple designed blocks is trained with time embedding $\mathbf{t}$ and class embedding $\mathbf{c}$, where a self-attention and a feed-forward module are involved in each block.
Note that they apply patchification on latent code $\mathbf{z}$ to extract a sequence of patch embeddings and depatchification operators are used to predict the denoised latent code $\mathbf{z}$.

\noindent\textbf{Revisit State Space Models (SSMs).}
State Space Models (SSMs)~\cite{gu2023mamba,gu2022efficiently} are pivotal in our approach due to their efficiency in handling sequences. 
Inspired by continuous systems, Mamba~\cite{gu2023mamba} models map a 1-D function or sequence $x(t) \in \mathbb{R}$ to $y(t) \in \mathbb{R}$, mediated through a hidden state $h(t) \in \mathbb{R}^\mathtt{N}$. The state dynamics are governed by the system matrices $\mathbf{A} \in \mathbb{R}^{\mathtt{N} \times \mathtt{N}}$, $\mathbf{B} \in \mathbb{R}^{\mathtt{N} \times 1}$, and $\mathbf{C} \in \mathbb{R}^{1 \times \mathtt{N}}$:
\begin{equation}
h'(t) = \mathbf{A}h(t) + \mathbf{B}x(t), \quad y(t) = \mathbf{C}h(t).
\end{equation}
For practical applications, Mamba~\cite{gu2023mamba} employs discrete approximations of these continuous systems, facilitated by a transformation parameter $\mathbf{\Delta}$. 
The continuous-to-discrete transformation commonly uses zero-order hold techniques:
\begin{equation}
\mathbf{\overline{A}} = \exp(\mathbf{\Delta} \mathbf{A}), \quad \mathbf{\overline{B}} = (\mathbf{\Delta} \mathbf{A})^{-1} (\exp(\mathbf{\Delta} \mathbf{A}) - \mathbf{I}) \mathbf{\Delta} \mathbf{B}.
\end{equation}
With these transformations, the system equations are discretized, allowing us to compute outputs at specific time steps:
\begin{equation}
h_t = \mathbf{\overline{A}}h_{t-1} + \mathbf{\overline{B}}x_{t}, \quad y_t = \mathbf{C}h_t.
\end{equation}
Finally, the output is computed using a structured global convolution, which processes the sequence through a convolutional kernel derived from the state transition matrices:
\begin{equation}
\mathbf{\overline{K}} = (\mathbf{C}\mathbf{\overline{B}}, \mathbf{C}\mathbf{\overline{A}}\mathbf{\overline{B}}, \dots, \mathbf{C}\mathbf{\overline{A}}^{L-1}\mathbf{\overline{B}}), \quad \mathbf{y} = \mathbf{z} * \mathbf{\overline{K}},
\end{equation}
where $L$ denotes the length of the input sequence $\mathbf{z}$, and $\overline{\mathbf{K}}$ represents a structured convolutional kernel, facilitating efficient sequence processing.

Although SSMs achieved promising performance on long sequence modeling, we empirically observe that the direct extension of Mamba~\cite{gu2023mamba} on point clouds does not work.
To address this problem, we propose a novel diffusion mamba architecture for 3D shape generation that can
effectively achieve the denoising processes on voxelized point clouds, as illustrated in Figure~\ref{fig: main_img}.

\subsection{Diffusion Mamba for 3D Shape Generation}\label{sec:diffmam}

Our novel DiM-3D framework synergizes the efficiency of Mamba’s linear computational architecture with the generative capabilities of diffusion models, specifically tailored for 3D shape generation. This integration allows for the effective management of spatial and temporal dimensions in 3D point clouds, crucial for capturing complex geometrical details in generated shapes.

The DiM-3D module repurposes the traditional Mamba model to accommodate 3D voxel data efficiently. By transforming voxelized point clouds $\mathbf{x} \in \mathbb{R}^{X \times Y \times Z \times C}$ into a sequence of flattened patches, the model optimizes the processing of volumetric data. Each patch $\mathbf{x}_p$ is a vector in $\mathbb{R}^{L \times (P^3 \cdot C)}$, where $P$ is the patch size, $C$ is the number of channels, and $L$ denotes the sequence length. Patches are then linearly projected to a higher dimensionality $D$ and augmented with positional embeddings. A class token is integrated to encapsulate global features of the entire shape. These token-level patch embeddings undergo multiple transformations in the DiM encoder, enhancing each patch’s feature representation through successive layers, enabling the generation of highly detailed and coherent 3D point clouds.

\subsection{DiM-3D Block with Linear Complexity}\label{sec:dim}

The DiM-3D block, a pivotal element of our architecture, employs bidirectional sequence modeling to efficiently process 3D point cloud data. This method is crucial for synthesizing intricate and high-resolution 3D shapes, as it allows for dynamic integration of global shape features throughout the generative process. Utilizing discrete state space models, the block adapts to the 3D spatial complexities, optimizing the handling of volumetric data.

\begin{table}[t]
	\renewcommand\tabcolsep{6.0pt}
    \renewcommand{\arraystretch}{1.1}
	\centering
 \caption{{\bf Detailed configurations of DiM-3D Models.} All models for the Small (S), Base (B), Large (L) and XLarge (XL) settings have comparable parameters to DiT-3D~\cite{mo2023dit3d} counterparts.}
 \label{tab: model_cfg}
	\scalebox{0.98}{
		\begin{tabular}{l|cc}
		\toprule
Method  & Layers $N$ & Hidden Size $d$\\
  \midrule
DiM-3D-S  & 16         & 384    \\
DiM-3D-B  & 16         & 768    \\
DiM-3D-L  & 32         & 1024   \\
DiM-3D-XL & 36         & 1152  \\
\bottomrule
			\end{tabular}}
\end{table}

Each DiM block starts with the normalization of the input token sequence, which is subsequently projected linearly to form the initial state vector $\mathbf{z}$. This vector undergoes transformations in both forward and backward directions, as shown in Figure~\ref{fig: main_img}.
In each direction, a 1-D convolution adjusts $\mathbf{z}$, resulting in intermediate state vectors $\mathbf{z}'$. These vectors are employed to compute transformation parameters $\mathbf{B}$, $\mathbf{C}$, and $\mathbf{\Delta}$, which are crucial for the bidirectional integration of data. The outputs from both directions are merged to ensure a cohesive output sequence, effectively leveraging information from all parts of the data structure.

This bidirectional processing mechanism enhances the system’s capability to comprehend and reconstruct spatial data, a significant advancement over traditional models designed primarily for 1-D sequences. The flexibility of the DiM blocks to adjust to various patch sizes and model dimensions allows for scalable and adaptable configurations. The model supports patch dimensions of 2, 4, 8, with complexity options ranging from Small (S) to XLarge (XL). This scalability is essential for accommodating diverse requirements in detail and resolution across different 3D modeling applications, aligning with the configurations similar to those used in the DiT model. Detailed model configurations are documented in Table~\ref{tab: model_cfg}, providing a comprehensive blueprint for implementation.

\section{Experiments}

\subsection{Experimental Setup}

\noindent\textbf{Datasets.}
We employ the ShapeNet dataset, focusing on the Chair, Airplane, and Car categories, consistent with benchmarks used in previous works. Each 3D shape in our experiments is represented by 2,048 points sampled from a set of 5,000 points, following the preprocessing protocol established in PointFlow~\cite{yang2019pointflow}. This normalization is applied globally across the entire dataset to maintain consistency.

\noindent\textbf{Evaluation Metrics.}
Our evaluation employs Chamfer Distance (CD) and Earth Mover’s Distance (EMD) to calculate 1-Nearest Neighbor Accuracy (1-NNA) and Coverage (COV). These metrics are crucial for assessing the generative quality of our models, where 1-NNA provides insights into the performance and diversity of generated point clouds, and COV measures the diversity of the generated shapes. Lower 1-NNA scores and higher COV scores indicate better performance.

\noindent\textbf{Implementation.}
DiM-3D is implemented in the PyTorch~\cite{paszke2019PyTorch} framework. We use a voxel input size of $32 \times 32 \times 32 \times 3$ and train our models for 10,000 epochs using the Adam optimizer with a learning rate of $1e-4$. The training employs a batch size of 128, and we set the total number of diffusion steps, $T$, to 1,000. 
In our standard configuration, the model size is S/4 with a patch size of 4.

\begin{table*}[t]
	\renewcommand\tabcolsep{2.0pt}
    \renewcommand{\arraystretch}{1.2}
	\centering
    \caption{{\bf Comparison results (\%) on shape metrics of our DiM-3D and state-of-the-art models.}
Our method significantly outperforms previous baselines in terms of all classes.}
 \label{tab: exp_sota}
	\scalebox{0.96}{
		\begin{tabular}{l|llll|llll|llll}
		\toprule
\multirow{3}{*}{Method} & \multicolumn{4}{c|}{Chair} & \multicolumn{4}{c|}{Airplane} & \multicolumn{4}{c}{Car} \\
& \multicolumn{2}{c}{1-NNA ($\downarrow$)} & \multicolumn{2}{c|}{COV ($\uparrow$)} & \multicolumn{2}{c}{1-NNA ($\downarrow$)} & \multicolumn{2}{c|}{COV ($\uparrow$)} & \multicolumn{2}{c}{1-NNA ($\downarrow$)} & \multicolumn{2}{c}{COV ($\uparrow$)}\\
& CD  & EMD & CD & EMD & CD & EMD & CD & EMD & CD & EMD & CD & EMD \\
  \midrule
  r-GAN~\cite{achlioptas2018learning}  & 83.69	& 99.70	& 24.27 & 15.13 & 98.40 & 96.79 & 30.12 & 14.32 & 94.46 & 99.01 & 19.03 & 6.539 \\
l-GAN (CD)~\cite{achlioptas2018learning} & 68.58	& 83.84	& 41.99 & 29.31 & 87.30 & 93.95 & 38.52 & 21.23 & 66.49 & 88.78 & 38.92 & 23.58 \\
l-GAN (EMD)~\cite{achlioptas2018learning} & 71.90	& 64.65	& 38.07 & 44.86 & 89.49 & 76.91 & 38.27 & 38.52 & 71.16 & 66.19 & 37.78 & 45.17 \\
PointFlow~\cite{yang2019pointflow} & 62.84	& 60.57	& 42.90 & 50.00 & 75.68 & 70.74 & 47.90 & 46.41 & 58.10 & 56.25 & 46.88 & 50.00 \\
SoftFlow~\cite{Kim2020SoftFlowPF}  & 59.21	& 60.05	& 41.39 & 47.43 & 76.05 & 65.80 & 46.91 & 47.90 & 64.77 & 60.09 & 42.90 & 44.60 \\
SetVAE~\cite{Kim2021SetVAE}  & 58.84	& 60.57	& 46.83 & 44.26 & 76.54 & 67.65 & 43.70 & 48.40 & 59.94 & 59.94 & 49.15 & 46.59 \\
DPF-Net~\cite{Klokov2020dpfnet} & 62.00	& 58.53	& 44.71 & 48.79 & 75.18 & 65.55 & 46.17 & 48.89 & 62.35 & 54.48 & 45.74 & 49.43 \\ 
\midrule
  DPM~\cite{luo2021dpm} & 60.05 & 74.77 & 44.86 & 35.50 & 76.42 & 86.91 & 48.64 & 33.83 & 68.89 & 79.97 & 44.03 & 34.94 \\
  PVD~\cite{zhou2021pvd} & 57.09 & 60.87 & 36.68 & 49.24 & 73.82 & 64.81 & 48.88 & 52.09 & 54.55 & 53.83 & 41.19 & 50.56 \\
LION~\cite{zeng2022lion} & 53.70 & 52.34 & 48.94 & 52.11 & 67.41 & 61.23 & 47.16 & 49.63 & 53.41 & 51.14 & 50.00 & 56.53 \\  
\midrule
GET3D~\cite{gao2022get3d} & 75.26 & 72.49 & 43.36 & 42.77 & -- & -- & -- & -- & 75.26 & 72.49 & 15.04 & 18.38 \\
MeshDiffusion~\cite{liu2023meshdiffusion} & 53.69 & 57.63 & 46.00 & 46.71 & 66.44 & 76.26 & 47.34 & 42.15 & 81.43 & 87.84 & 34.07 & 25.85 \\  
\midrule
DiT-3D-XL~\cite{mo2023dit3d} & 49.11 & 50.73 & 52.45 &  54.32 & 62.35 & 58.67 & 53.16 & 54.39 & 48.24 & 49.35 & 50.00 & 56.38 \\ 
FastDiT-3D-S~\cite{mo2023fastdit3d} & 50.35 & 50.27  & 58.53 & 60.79 &  61.83 &  57.86 &  58.21 & 58.75 & 47.81 & 48.83 & 53.86 & 59.62\\ 
DiM-3D (ours) & \bf 45.78 & \bf 47.07 & \bf 57.89 & \bf 55.61 & \bf 59.72 & \bf 55.38 & \bf 60.67 & \bf 61.25 & \bf 45.79 & \bf 46.28 & \bf 59.63 & \bf 65.78 \\
\bottomrule
			\end{tabular}}
\end{table*}

\subsection{Comparison to prior work}

In this work, we propose a novel and effective diffusion transformer for 3D shape generation.
In order to validate the effectiveness of the proposed DiM-3D, we comprehensively
compare it to previous non-DDPM, DDPM, and Diffusion Transformer-based approaches.

Our DiM-3D is evaluated against a spectrum of established models to underline its efficacy in 3D shape generation. We benchmark against a range of methods from traditional GANs like r-GAN and 1-GAN to advanced DDPM approaches like DiT-3D and recent innovations like MeshDiffusion. Each of these models represents a unique approach to 3D shape generation, using different architectural insights and learning mechanisms.
Our method particularly excels against the state-of-the-art DiT-3D, showing significant improvements in both 1-NNA and COV metrics across all categories. Notably, DiM-3D achieves better precision and diversity in generating detailed point clouds, particularly for complex objects like airplanes and cars. These results are indicative of the effectiveness of our Mamba-based diffusion approach in handling high-resolution, volumetric data efficiently.

The quantitative outcomes in Table~\ref{tab: exp_sota}, highlight DiM-3D’s superior performance over contemporary methods, demonstrating the advantages of integrating Mamba’s linear complexity handling into diffusion-based generative modeling. This integration not only enhances the quality of the generated shapes but also significantly boosts the diversity of the output, confirming the effectiveness of our model architecture and training regimen.

\begin{figure*}[t]
\centering
\includegraphics[width=0.98\linewidth]{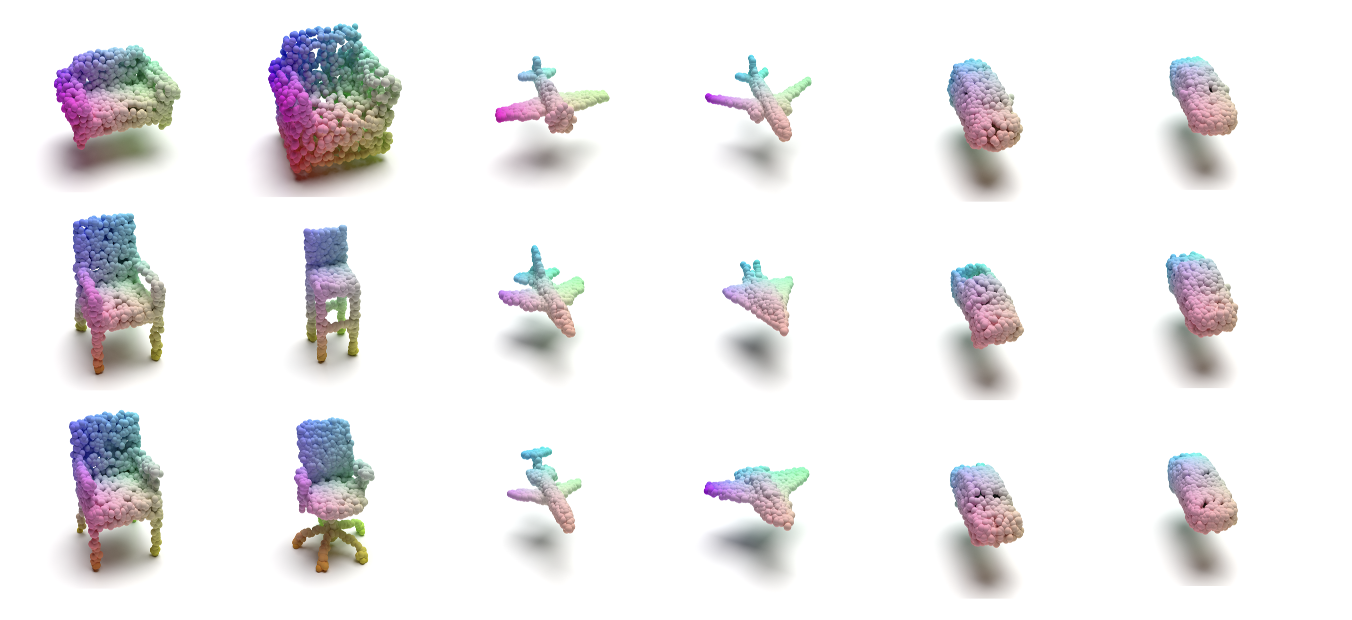}
\caption{{\bf Qualitative visualizations of generated 3D point clouds.}
Our DiM-3D achieves high-fidelity and diverse 3D point cloud generation across different categories.
}
\label{fig: exp_vis}
\end{figure*}

The visual results of our experiments, shown in Figure~\ref{fig: exp_vis}, provide a direct comparison of the generative capabilities of DiM-3D against other leading methods. These visualizations clearly depict the refined and realistic nature of the point clouds generated by our model, showcasing the practical effectiveness of applying the diffusion transformer framework to 3D shape generation.
Through these experiments, we validate the claims made in the introduction and abstract, establishing DiM-3D as a leading architecture in the field of 3D shape generation. The empirical evidence supports our model’s capacity to set new benchmarks for fidelity and diversity in the generation of complex 3D shapes.

\subsection{Experimental analysis}

In this section, we performed ablation studies to demonstrate the efficiency of introducing mamba on point cloud generation.
We also conducted extensive experiments to explore the scalability of conditional generation and large-scale training in more classes and different model sizes.

\begin{table}[t]
	\renewcommand\tabcolsep{6.0pt}
    \renewcommand{\arraystretch}{1.1}
	\centering
 \caption{{\bf Ablation results on Gflops of DiM-3D-XL/2 models versus DiT-3D-XL/2 models across diverse voxel sizes.} 
 All Gflops are calculated by thop package for a fair comparison.}
 \label{tab: ab_flops}
	\scalebox{0.98}{
		\begin{tabular}{l|cc}
		\toprule
Method  & Voxel Size & Model Gflops ($\downarrow$)  \\
  \midrule
DiT-3D-XL/2 & 256x256x256        & 343.28                  \\
DiM-3D-XL/2 & 256x256x256        & \bf 294.58                  \\ \hline
DiT-3D-XL/2 & 512x512x512        & 1371.08                 \\
DiM-3D-XL/2 & 512x512x512        & \bf 1175.7                  \\ \hline
DiT-3D-XL/2 & 1024x1024x1024       & 5482.28                 \\
DiM-3D-XL/2 & 1024x1024x1024       & \bf 4700.17                 \\ \hline
DiT-3D-XL/2 & 2048x2048x2048       & OOM \\
DiM-3D-XL/2 & 2048x2048x2048       & \bf 18798.04  \\       
\bottomrule
			\end{tabular}}
\end{table}

\noindent\textbf{Efficiency of Mamba on Gflops.}
We quantitatively assessed the computational efficiency of the Mamba architecture by measuring the Giga Floating Point Operations (Gflops) required for processing using the thop~\footnote{\url{https://github.com/Lyken17/pytorch-OpCounter}} package package. Our findings, summarized in Table~\ref{tab: ab_flops}, indicate that integrating the Mamba architecture significantly reduces the computational load compared to traditional diffusion models. This efficiency stems from the linear complexity of the Mamba framework, which scales favorably with increased input sizes and model complexity.

\begin{table}[t]
	\renewcommand\tabcolsep{6.0pt}
    \renewcommand{\arraystretch}{1.1}
	\centering
 \caption{{\bf Comparison results on conditional generation for point cloud completion.} All reported results are averaged on three different running seeds. Our DiM-3D achieves the best performance.}
 \label{tab: ab_condition}
	\scalebox{0.98}{
		\begin{tabular}{l|ll|ll|ll}
           \toprule
\multirow{2}{*}{Method} & \multicolumn{2}{c|}{Chair}                             & \multicolumn{2}{c|}{Airplane}                           & \multicolumn{2}{c}{Car}                               \\
                        & CD($\downarrow$)                        & EMD($\downarrow$)                       & CD($\downarrow$)                         & EMD($\downarrow$)                       & CD($\downarrow$)                        & EMD($\downarrow$)                       \\ \midrule
PVD~\cite{zhou2021pvd}                     & 3.211 & 2.939 & 0.4415 & 1.030 & 1.774 & 2.146 \\
LION~\cite{zeng2022lion}                    & 2.725                     & 2.863                     & 0.4035                     & 0.9732                    & 1.405                     & 1.982                     \\
DiT-3D~\cite{mo2023dit3d}                 & 2.216                     & 2.385                     & 0.3521                     & 0.9235                    & 1.126                     & 1.513                   \\ 
DiM-3D (ours) & \bf 1.786 & \bf 1.763 & \bf 0.2516 & \bf 0.7859 & \bf 0.832 & \bf 0.956 \\ \bottomrule 
\end{tabular}}
\end{table}

\noindent\textbf{Conditional generation.}
To examine the capability of DiM-3D in conditional generation scenarios, we conducted point cloud completion experiments on the ShapeNet dataset. We compared our model's performance against established methods like PVD and LION using Chamfer Distance and Earth Mover’s Distance as metrics. The results in Table~\ref{tab: ab_condition} show that DiM-3D consistently outperforms these benchmarks across all categories (Airplane, Chair, Car). This superior performance underscores the model's effectiveness in handling conditional inputs and generating detailed, accurate 3D reconstructions.

\begin{table}[t]
	\renewcommand\tabcolsep{6.0pt}
    \renewcommand{\arraystretch}{1.1}
	\centering
 \caption{{\bf Comparison results on large-scale generation for more classes.} All reported models are trained on ShapeNet-55. Our DiM-3D achieves the best results.}
 \label{tab: ab_class}
	\scalebox{0.98}{
		\begin{tabular}{l|cc|cc}
           \toprule
\multirow{2}{*}{Method} & \multicolumn{2}{c|}{Mug}                             & \multicolumn{2}{c}{Bottle}                           \\
                 & 1-NNA CD($\downarrow$) & COV-CD($\uparrow$) & 1-NNA CD($\downarrow$) & COV-CD($\uparrow$)   \\                      \midrule
LION~\cite{zeng2022lion} & 70.45 & 31.82 & 61.63 & 39.53 \\
Point-E~\cite{nichol2022pointe} & 65.73 & 36.78 & 58.16 & 43.72 \\
DiT-3D~\cite{mo2023dit3d} & 57.39 & 45.26 & 53.26 & 51.28 \\
DiM-3D (ours) & \bf 52.16 & \bf 51.78 & \bf 50.23 & \bf 58.76 \\ \bottomrule 
\end{tabular}}
\end{table}

\noindent\textbf{Scaling to more classes.}
Beyond the single-class comparison, we trained our DiT-3D on the large-scale ShapeNet-55 dataset with 55 diverse classes covering vehicles, furniture, and daily necessities. We compare the newly trained model with the state-of-the-art point cloud generation model, LION~\cite{zeng2022lion}, on Mug and Bottle generation in Table~\ref{tab: ab_class}. 
Our method achieves the best results in terms of all metrics.
We also conducted additional experiments comparing our DiM-3D model with Point-E~\cite{nichol2022pointe}, wherein we utilized the class as a text prompt. 
The training was performed on the large-scale ShapeNet-55~\cite{chang2015shapenet} dataset, which comprises 55 diverse classes encompassing vehicles, furniture, and daily necessities. 
Specifically, we evaluated the performance of Mug and Bottle generation, and the results are presented in the Table below. Our method consistently outperforms Point-E~\cite{nichol2022pointe} across all metrics, highlighting the superiority of our approach.

\begin{table*}[t]
	\renewcommand\tabcolsep{2.0pt}
    \renewcommand{\arraystretch}{1.2}
	\centering
    \caption{{\bf Ablation results (\%) on shape metrics of our DiM-3D models.}
Our method scales well in large-sized parameters in terms of all metrics.}
 \label{tab: ab_model}
	\scalebox{0.96}{
		\begin{tabular}{l|llll}
		\toprule
\multirow{3}{*}{Model} & \multicolumn{4}{c}{Chair} \\
& \multicolumn{2}{c}{1-NNA ($\downarrow$)} & \multicolumn{2}{c}{COV ($\uparrow$)} \\
& CD  & EMD & CD & EMD \\
  \midrule
DiM-3D-S & 57.27 & 54.10 & 51.84 & 50.83 \\
DiM-3D-B & 55.41 & 52.18 & 52.03 & 51.49 \\
DiM-3D-L & 48.71 & 51.39 & 54.12 & 52.33 \\
DiM-3D-XL & \bf 45.78 & \bf 47.07 & \bf 57.89 & \bf 55.61 \\
\bottomrule
			\end{tabular}}
\end{table*}

\noindent\textbf{Scaling model sizes.}
Finally, we explored the impact of varying model sizes on performance by scaling DiM-3D from S/4 to XL/4. The models were trained for an extended duration of 2,000 epochs to stabilize learning across sizes. The results, presented in Table~\ref{tab: ab_model}, indicate consistent performance improvements with increasing model size. This scalability is a testament to the adaptability of the DiM-3D architecture, confirming its capability to generate high-fidelity and diverse 3D shapes across different configurations.
Through these experimental analyses, we validate the practical efficacy of DiM-3D. Our findings demonstrate not only the theoretical advantages of our model but also its operational superiority in diverse and demanding generative tasks in 3D modeling.

\section{Conclusion}

In this work, we present DiM-3D, a novel diffusion mamba architecture tailored for the efficient and scalable generation of high-fidelity 3D point clouds. By integrating the selective state space approach of the Mamba architecture, our model overcomes the computational inefficiencies of traditional diffusion models, maintaining linear complexity with respect to sequence length while achieving significant improvements in inference speed and reduction in computational demands.
Our comprehensive evaluations on the ShapeNet dataset, encompassing categories such as Chairs, Airplanes, and Cars, have demonstrated that DiM-3D sets new benchmarks in the field of 3D shape generation. Through extensive experiments, including comparisons with state-of-the-art models and a series of ablation studies, DiM-3D has shown superior performance in generating detailed and diverse 3D shapes. Notably, our model excels in conditional generation tasks and scales effectively with increased model sizes and the number of classes, indicating robust adaptability and high scalability.
The results of our ablation studies further underscore the efficiency of the Mamba architecture in reducing Gflops and enhancing the generative capabilities of our diffusion model. The model's proficiency in managing larger datasets and various model configurations without compromising output quality illustrates its potential for broader applications in industrial design, virtual reality, and autonomous navigation systems.

\section*{Limitation}
While DiM-3D demonstrates substantial advancements in the field of 3D shape generation, certain limitations merit attention. Firstly, the model’s complexity and the large-scale data requirements might pose challenges in environments with limited computational resources. Despite the efficiency improvements introduced by the Mamba architecture, the handling of extremely high-resolution voxel data and large datasets still demands significant computational power, which could limit accessibility for researchers and practitioners with constrained resources.
Secondly, the model's dependency on the quality and diversity of the training data can affect its generalizability. While DiM-3D performs exceptionally on the ShapeNet dataset, its effectiveness on datasets with less variety or poorer annotations might not be as pronounced. This reliance on high-quality, extensive training data could restrict the model’s applicability to areas where such data are not readily available.

\section*{Broader Impact}
The development of DiM-3D has broader implications for both technology and society. In industries such as gaming, film, and virtual reality, the ability to generate realistic 3D environments and objects efficiently can significantly enhance user experiences and reduce production costs. Similarly, in autonomous systems, improved 3D modeling can enhance navigation capabilities and safety features by providing more accurate simulations and training environments.

\clearpage

\bibliography{reference}
\bibliographystyle{unsrt}




\end{document}